\documentclass[10pt,twocolumn,letterpaper]{article}

\usepackage{iccv}
\usepackage{times}
\usepackage{epsfig}
\usepackage{graphicx}
\usepackage{amsmath}
\usepackage{amssymb}
\usepackage{color}
\usepackage{caption}

\usepackage[ruled,vlined]{algorithm2e}
\usepackage{ctable}
\usepackage{caption}
\usepackage{enumitem}

\renewcommand{\paragraph}[1]{\vspace{1mm}\noindent\textbf{#1}}
\newcommand{\Tref}[1]{Table~\ref{#1}}

\newcommand{\fref}[1]{Fig.~\ref{#1}}

\usepackage[pagebackref=true,breaklinks=true,letterpaper=true,colorlinks,bookmarks=false]{hyperref}

\iccvfinalcopy 

\def\iccvPaperID{2855} 

\ificcvfinal\fi
\begin{document}

\title{Copy-and-Paste Networks for Deep Video Inpainting}

\author{Sungho Lee\\
Yonsei University\\
\and
Seoung Wug Oh\\
Yonsei University\\
\and
DaeYeun Won\\
Hyundai MNSOFT\\
\and
Seon Joo Kim\\
Yonsei University\\
}
\maketitle
\thispagestyle{empty}

\begin{abstract}
   We present a novel deep learning based algorithm for video inpainting.
   Video inpainting is a process of completing corrupted or missing regions in videos.
   Video inpainting has additional challenges compared to image inpainting due to the extra temporal information as well as the need for maintaining the temporal coherency.
   We propose a novel DNN-based framework called the Copy-and-Paste Networks for video inpainting that takes advantage of additional information in other frames of the video. 
   The network is trained to copy corresponding contents in reference frames and paste them to fill the holes in the target frame. 
   Our network also includes an alignment network that computes affine matrices between frames for the alignment, enabling the network to take information from more distant frames for robustness.
   Our method produces visually pleasing and temporally coherent results while running faster than the state-of-the-art optimization-based method.
   In addition, we extend our framework for enhancing  over/under exposed frames in videos. Using this enhancement technique, we were able to
   significantly improve the lane detection accuracy on road videos.
\end{abstract}


\section{Introduction} 
Inpainting is a task of completing an image that has empty pixels by filling the empty regions with visually plausible pixels.
Inpainting is very useful in image editing process, and is usually utilized to generate more satisfying images by removing unwanted objects in images.
There is a large body of literature on image inpainting and significant progress has been made recently by employing deep learning for image inpainting. 
Impressive inpainting results are reported by applying evolving deep generative models~\cite{goodfellow2014generative}, synthesizing visually pleasing images even for complex scenes. 

In this paper, we focus on the video inpainting problem. 
Videos with additional temporal information makes the already difficult problem even more challenging.
In addition to filling the holes for every frame, the algorithm has to ensure that the completed frames are temporally consistent. 
Due to these challenges, we have only seen one work that tackles the problem using deep neural networks (DNN)~\cite{Kim2019CVPR},
compared to the image inpainting problem where many deep learning based algorithms have been introduced.

\begin{figure}
\centering
\includegraphics[width=1.0\linewidth]{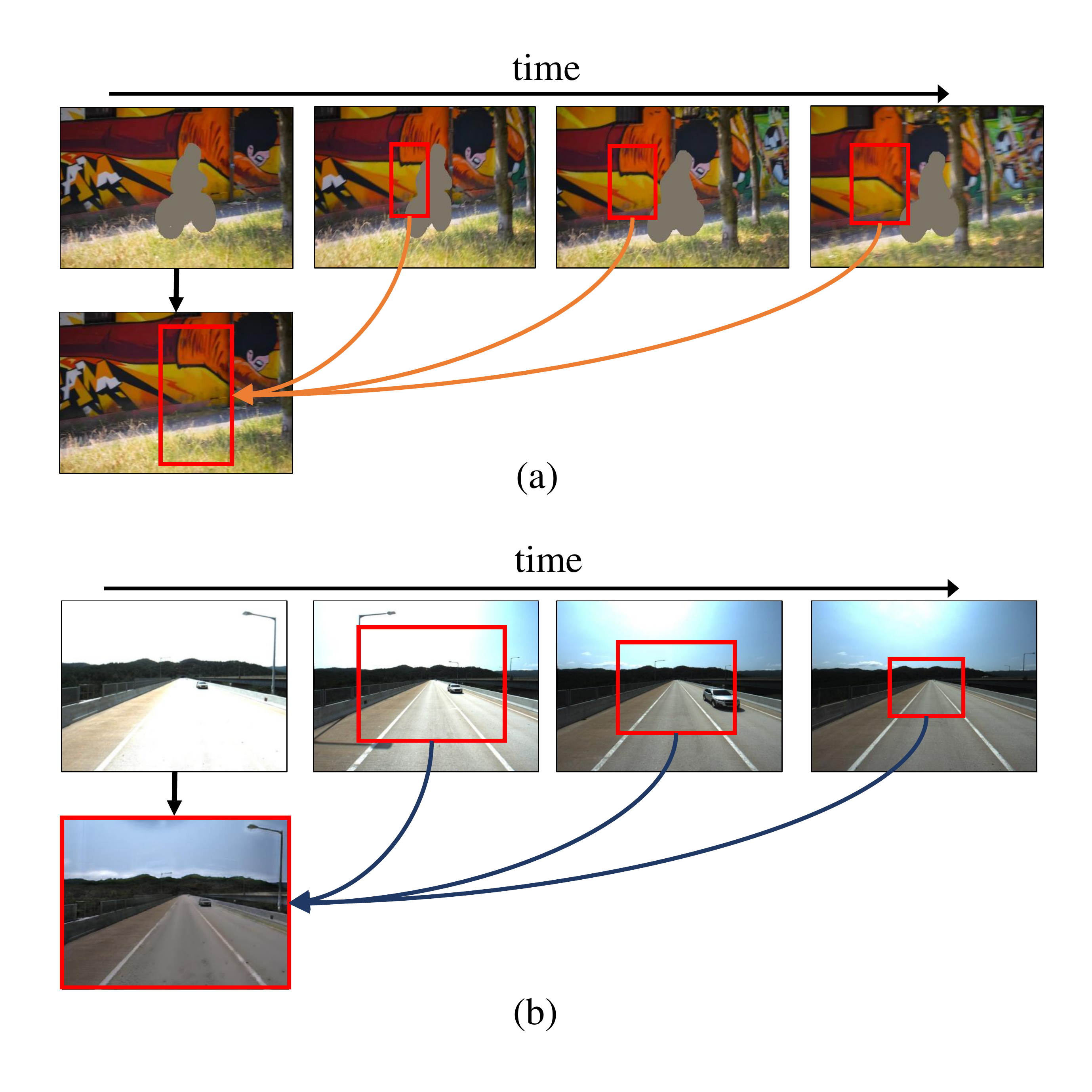}
\caption{(a) We propose a DNN framework for video inpainting. Our Copy-and-Paste network learns to find corresponding pixels in other frames to fill in the holes in the given frame. (b) Another application of our framework for restoring an over-saturated image. }
\label{fig:teaser}
\end{figure}


While video inpainting is more challenging compared to image inpainting, it inherently includes more cues for the problem as valid pixels for missing regions in a frame may exist in other frames.
Therefore, we propose a novel DNN based framework called the \textit{Copy-and-Paste Networks} for video inpainting that takes advantage of additional information in other frames in the video. 
As the name suggests, the network is trained to copy the necessary pixels from other frames and paste those pixels on the holes in the current frame (\fref{fig:teaser}).

The key components of our DNN system are the alignment and the context matching. 
To find corresponding pixels in other frames for the holes in the given frame, the frames need to be registered first. 
We propose a self-supervised alignment networks, which estimates affine matrices between frames. 
While DNNs for computing the affine matrix or homography exist~\cite{detone2016deep, Jaderberg2015NIPS, nguyen2018unsupervised}, our alignment method is able to deal with holes in images when computing the affine matrices.
After the alignment, the novel context matching algorithm is used to compute the similarity between the target frame and the reference frames.
The network learns which pixels are valuable for copying through the context matching, and those pixels are used to paste and complete an image.
By progressively updating the reference frames with the inpainted results at each step, the algorithm can produce videos with temporal consistency. 

Our results are comparable to the state-of-the-art method~\cite{Huang2016siggraph}, and outperform other deep learning based approaches~\cite{Kim2019CVPR,yu2018generative}.
Moreover, we can easily extend our method for restoring saturated/under-exposed images as shown in (\fref{fig:teaser}(b)).
By enhancing the saturated/under-exposed images, we were able to significantly increase the lane detection accuracy. 

In summary, the major contribution of our paper is as follows:
\begin{itemize}
    \item We propose a self-supervised deep alignment networks that can compute affine matrices between images that contain large holes.
    \item We propose a novel context-matching algorithm to combine reference frame features based on similarity between images.
    \item Our method produces visually pleasing completed videos, running much faster than the state-of-the-art method. Additionally, we extend
     our framework for enhancing over/under exposed frames in videos that can help to improve other vision tasks such as the lane detection.
\end{itemize}

\section{Related works}
\subsection{Image Inpainting}
In traditional image inpainting methods, an image is filled by referencing pixels outside the hole in the image or in the external image database.
As one of the most representative inpainting methods, PatchMatch~\cite{barnes2009patchmatch} reconstructs the missing region by searching the patches outside the hole based on the approximate nearest neighbor algorithm.
With this type of approach, however, it is difficult to inpaint images with complicated scenes, or when the images do not contain sufficient information for filling the holes.

Since deep image inpainting has been introduced in \cite{IizukaSIGGRAPH2017, PathakCVPR2016},
many deep generative models for image inpainting have been proposed recently, showing impressive restoration results on complex scenes.
Yu~\etal\cite{yu2018generative} proposed the contextual attention module between the completed structure of the hole area and the patches outside the hole. 
Liu~\etal\cite{Liu2018PartialConv} and Yu~\etal\cite{yu2018free} applied the partial convolution and the gated convolution to compensate the weakness of the vanilla convolution for image inpainting.
In particular, Liu~\etal\cite{Liu2018PartialConv} corrected the blurred results based on the perceptual and the style loss without the adversarial loss. 

\subsection{Video Inpainting}
Video inpainting has additional challenges of restoring the holes in every frame and maintaining the temporal consistency between reconstructed frames. 
Meanwhile, unlike in image inpainting, one can utilize redundant information between frames of video in video inpainting. 
However, directly exploiting the redundant information in videos is difficult due to image variation from the movements of the camera and the objects. 
To compensate for the movements, Granados~\etal\cite{Granados2012ECCV} proposed to align the frames based on the homographies. 
They also applied the optical flow between completed frames to maintain the temporal consistency. 


In \cite{Newson2014siam}, Newson~\etal proposed 3D PatchMatch to maintain the temporal consistency in addition to using the affine transformation to compensate the motion.
While the spatio-temporal patches improve the short-term temporal consistency, the long-term consistency of complicated scenes remained as a limitation.
To solve this limitation, Huang~\etal\cite{Huang2016siggraph} proposed the optical flow optimization in spatial patches to complete images while preserving the temporal consistency.
This method shows the state-of-the-art performance up until now. 
All the methods explained above are based on a heavy optimization, and therefore suffers in the computational time, limiting their practical use.

Wang~\etal\cite{Wang2019AAAI} proposed the first deep learning based video inpainting by using 3D encoder-decoder networks.
However, this work does not cover the object removal task in general videos, and was only applied to a few specific domains. Kim~\etal\cite{Kim2019CVPR} proposed 3D-2D encoder-decoder networks to complete the missing contents efficiently.
The temporal consistency is maintained through a recurrent feedback and a memory layer with the flow and the warping loss.
The temporal window for the referencing is small in their method, and therefore it is difficult to use valid pixels in distant frames, resulting in a limited performance for scenes with large objects or slowly moving objects. 
\begin{figure*}
\centering
\includegraphics[width=1.0\linewidth]{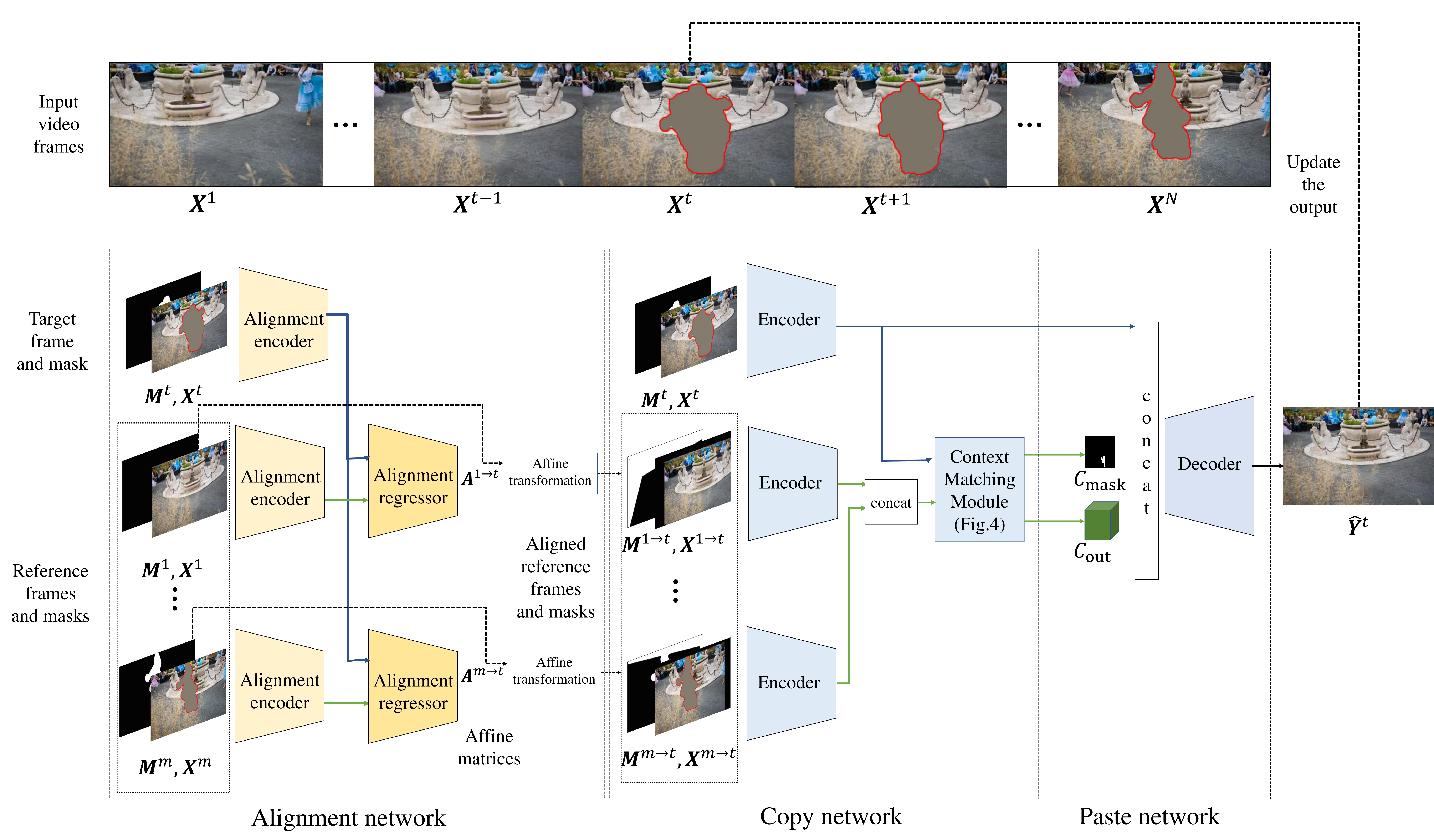}
\caption{Network Overview. Our framework consists of 3 sub-networks: alignment network, copy network, and paste network.}
\label{fig:networks}
\end{figure*}

Our copy-and-paste network overcome the issues in \cite{Kim2019CVPR} by aligning the frames with affine matrices computed by our alignment network instead of using the optical flow.
With the novel context matching algorithm, our method can extract valid pixels in distant frames, resulting in more accurate reconstruction for general scenes. 
The performance of our method is comparable to the state-of-the-art method in~ \cite{Huang2016siggraph} while being more practical with
faster runtime due to the feed forward nature of DNNs. 



\section{Copy-and-Paste Network Algorithm}
The overview of our framework is shown in \fref{fig:networks}. 
The system takes a video ($\boldsymbol{X}$) annotated with the missing pixels ($\boldsymbol{M}$) in each frame and outputs ($\boldsymbol{\hat{Y}}$) the completed video. 
The video is processed frame-by-frame in the temporal order.
We call the frame to be filled as the target frame and the other frames as the reference frames.
For each target frame, our network completes the missing region by copying-and-pasting contents from the reference frames.

To complete a target frame, each reference frame is first aligned to the target frame through the alignment network.
Then in the copy network, pixels to be copied from the aligned reference frames are determined by the context matching module. 
Finally, the outputs from the copy networks are decoded to produce inpainted target frame in the paste network. 
The input video in the memory is updated with the completed frame, which will subsequently be used as a reference frame, providing more information for the following frames. 
\begin{figure}
\centering
\includegraphics[width=1.0\linewidth]{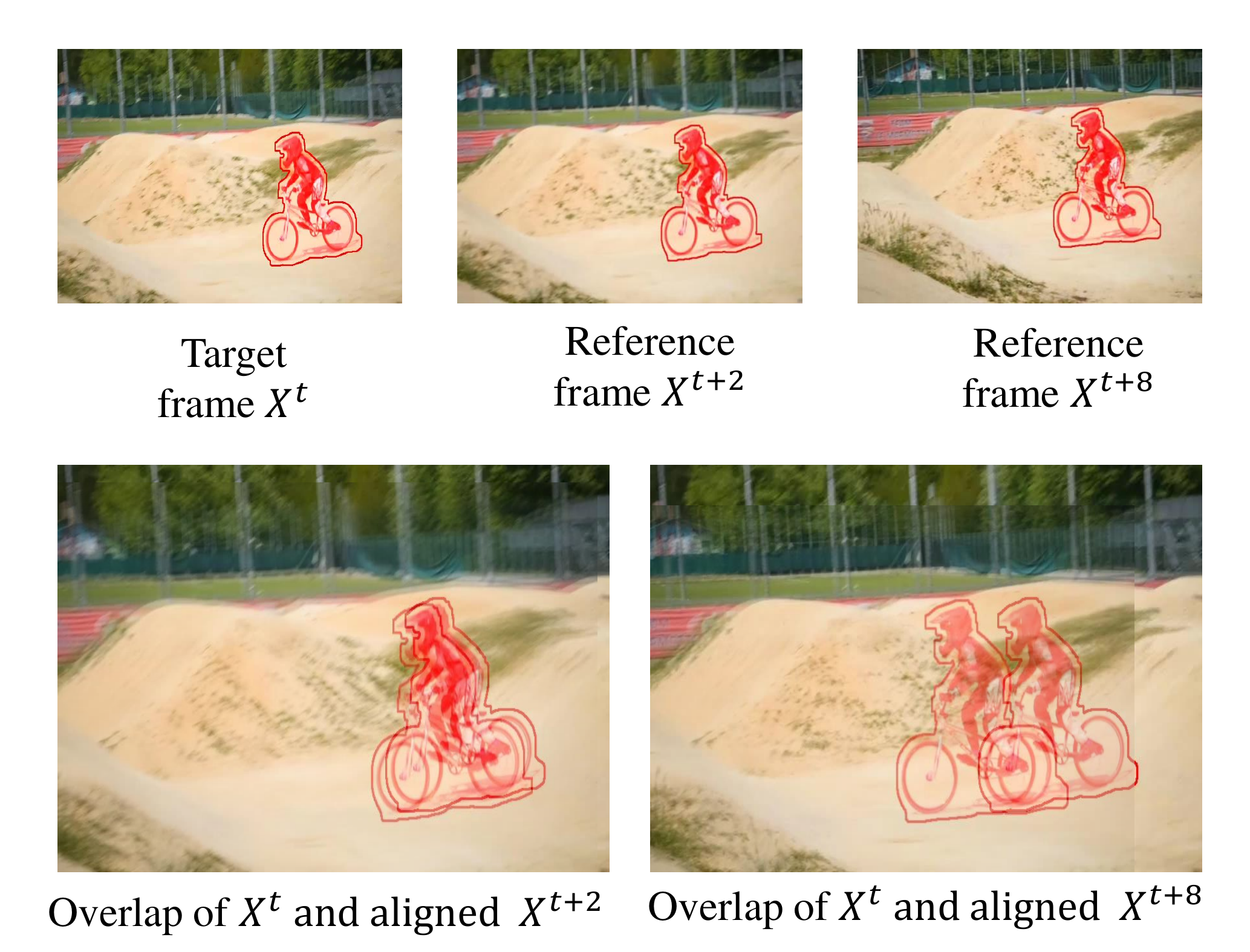}
\caption{Using affine transformation for the alignment yields larger temporal search compared to the optical flow based alignment. More distant reference frame provides more valuable information as the overlap of the hole regions is smaller.}
\label{fig:overlap}
\end{figure}

\subsection{Alignment Network}
In video inpainting, a large temporal window is essential as valuable information is more likely to be in distant frames. 
With an optical flow based alignment as used in ~\cite{Kim2019CVPR}, the temporal range of information is too small to extract useful information.
As illustrated in \fref{fig:overlap}, a reference frame temporally close to the target frame lacks information to fill the hole as there are too much overlap between the holes in the images.
Moreover, computing optical flows between images with holes is more difficult as the holes themselves become occlusion factors.
Therefore, our alignment network estimates the affine matrices to align the reference frames with the target frame. 

\begin{figure}
\centering
\includegraphics[width=1.0\linewidth]{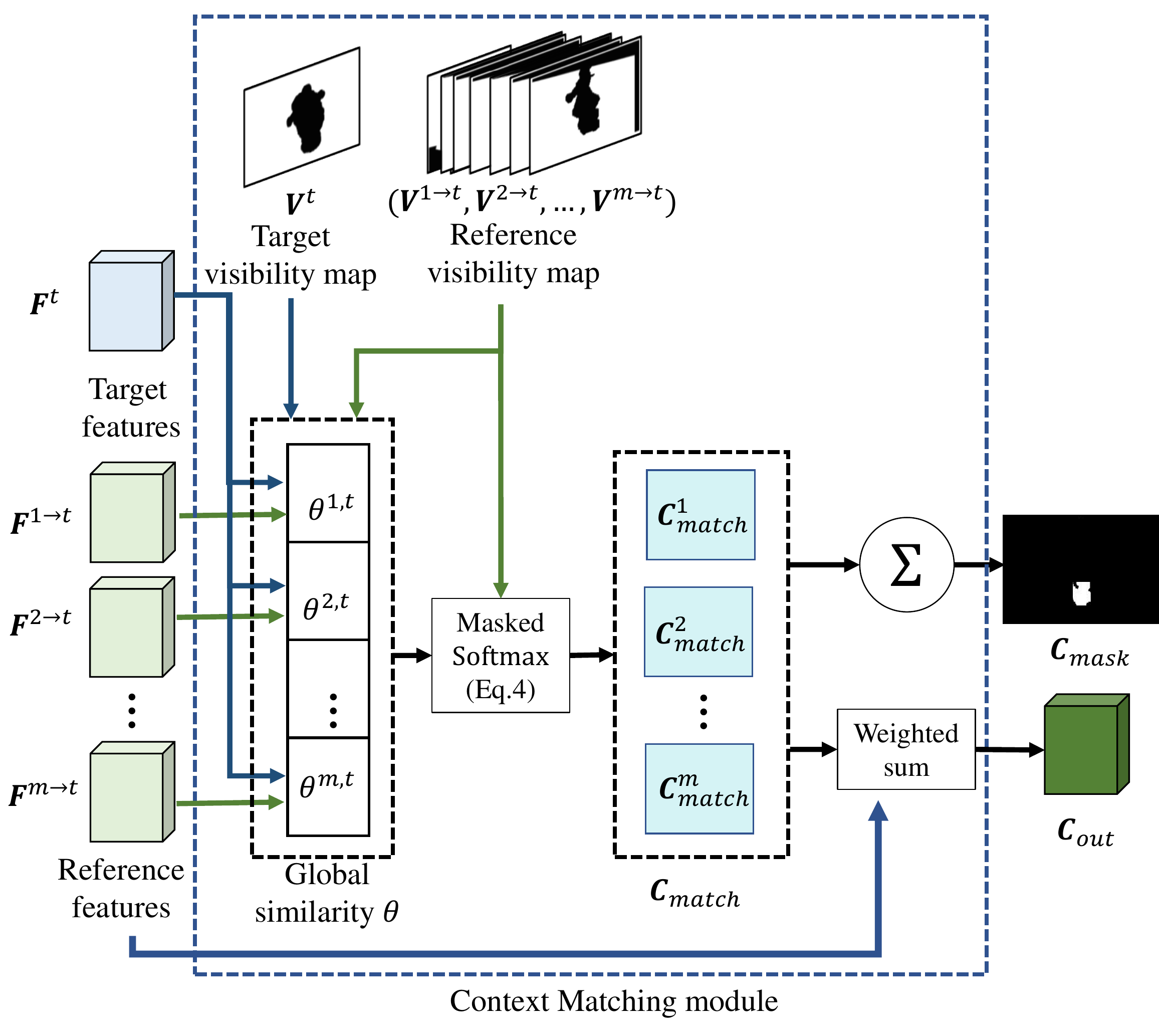}
\caption{Detailed illustration of the context matching module.}
\label{fig:contextmatching}
\end{figure}

The alignment network consists of shared alignment encoders and alignment regressors.
Details on the network architectures are provided in the supplementary materials.
To train the alignment network, we minimize the self-supervised loss, which is the L1 distance between the target frame ($\boldsymbol{X}^t$) and the aligned reference frame ($\boldsymbol{X}^{r\to t}$).
To exclude the hole regions, this pixel-wise loss is only measured with pixels that are valid in both images as follows:  
\begin{equation} \label{eq:align}
\begin{aligned}
    \mathcal{L}_\text{align} &= {\sum_{r} {||} \boldsymbol{V} \odot ({\boldsymbol{X}^t }-{\boldsymbol{X}^{r\to t}}) {||}_1,}
\end{aligned}
\end{equation}
where $\boldsymbol{V} = \boldsymbol{V}^t \odot \boldsymbol{V}^{r \to t} $ is the visibility map, $\odot$ is the element-wise product, $t$ is the target frame index, and $r$ is the reference frame index\footnote{The symbol $r\to t$ indicates aligning a reference frame $r$ to a target frame $t$. $\boldsymbol{V}^{r\to t}$ indicates the visibility map of the reference aligned to the target}.
The visibility map is computed from the given masks, where 0 indicates hole pixels and 1 represents non-hole pixels.
\begin{figure}
\centering
\includegraphics[width=1.0\linewidth]{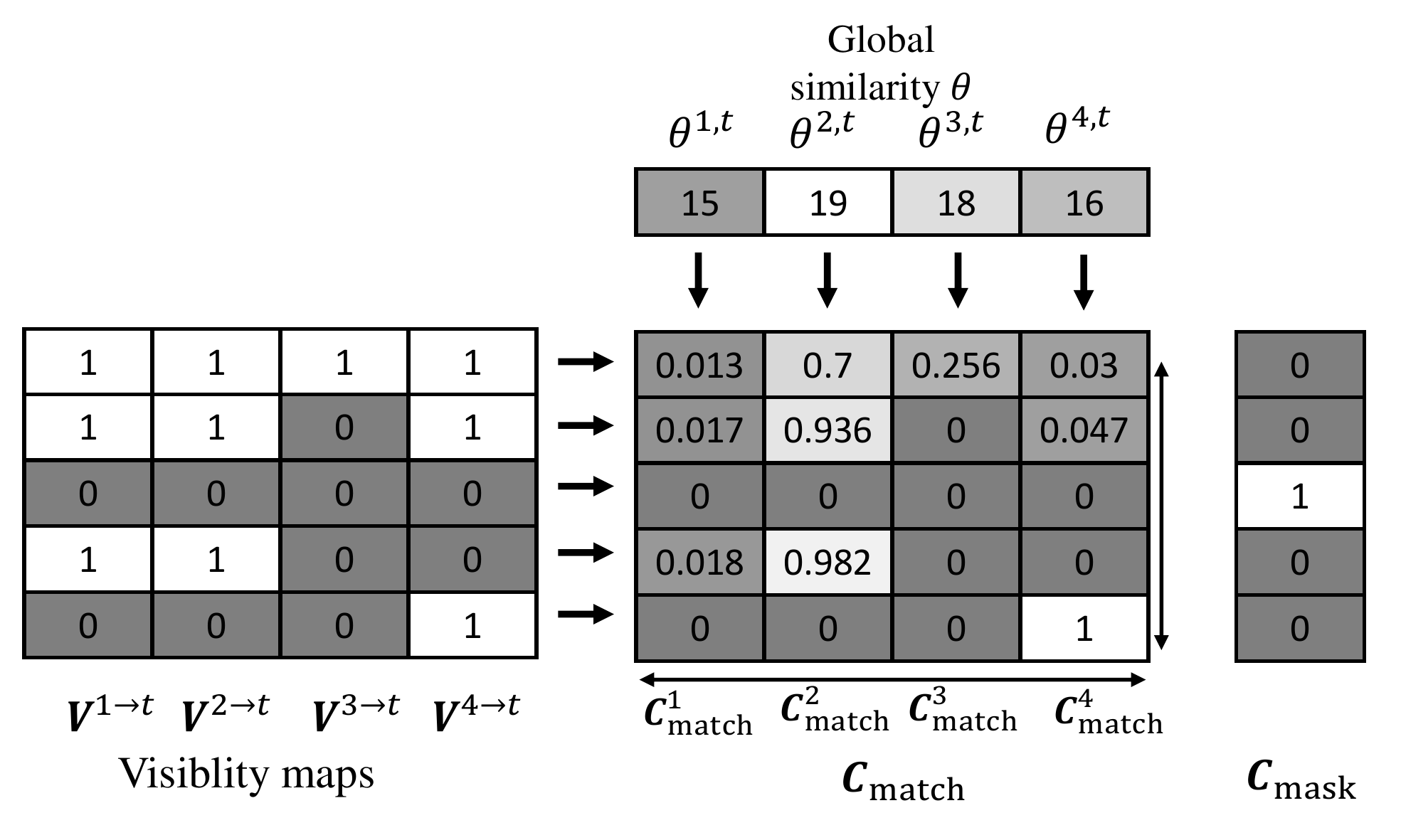}
\caption{{An 1-D example of masked softmax.}}
\label{fig:maskedsoftmax}
\end{figure}
Note that the alignment network is jointly trained with other networks in an end-to-end manner, not independently.

\subsection{Copy-and-Paste Network}
After the frame alignment, the aligned frames are mapped into the feature space through the shared encoders.
The context matching module computes the importance of each pixel in the reference frames in completing the holes as well as a mask ($\boldsymbol{C}_\text{mask}$) indicating the visibility of each pixel throughout the video. 
Finally, the decoder takes the output of the context matching module in addition to the target frame feature to restore values for the missing pixels. 

\paragraph{Encoder}  
Encoder networks extract the features from the target and the aligned reference frames. 
The input to the encoder is a concatenation of an RGB image and the corresponding binary mask.
The details on the architecture will be described in the supplementary materials.

\paragraph{Context matching module} 
Together with the encoder, the context matching module constitutes the copy network.
The context matching module is illustrated in \fref{fig:contextmatching}.
First, global similarities ($\theta^{r,t}$)  between the aligned reference frames and the target frame in the feature space is computed as follows:

\begin{equation}
    \theta^{r,t} = {1 \over{\sum_{(x,y)} \boldsymbol{V}(x,y)}} \cdot \sum_{(x,y)}\boldsymbol{V}( x,y)\cdot \boldsymbol{F}^{t}(x,y) \cdot \boldsymbol{F}^{r\to t}(x,y).
\end{equation}
The above equation is basically computing the cosine similarity between the two feature maps, excluding the hole pixels. 

Then, a saliency map $\boldsymbol{C}^{r}_\text{match}$ for each reference frame is computed as follows:
\begin{equation}
    \boldsymbol{S}^{r,t} = \theta^{r,t} \cdot \boldsymbol{V}^{r\to t},
    \label{eq:context_matcher1}
\end{equation}
\begin{equation}
    \boldsymbol{C}^{r}_\text{match}(x,y)= 
    \begin{cases}{{\large{exp}({\boldsymbol{S}^{r,t}(x,y)}})\over{\sum_{r}\large{exp}({\boldsymbol{S}^{r,t}(x,y)}})}&\text{if}\, \boldsymbol{V}^{r\to t}(x,y) = 1\\
    0 & otherwise.
    \end{cases}
\end{equation}

\fref{fig:maskedsoftmax} simplifies the steps for computing the saliency map in 1-D.
Each pixel value in the saliency map $\boldsymbol{C}^{r}_\text{match}$ holds the weight that specific pixels have on filling the hole in the target. 
The reference features are aggregated through a weighted sum with the $\boldsymbol{C}^{r}_\text{match}$, producing the features to be used for the decoder ($C_\text{out}$).
\begin{equation}
    \boldsymbol{C}_\text{out}(x,y) =
    \sum_{r} {\boldsymbol{F}^{r\to t}(x,y)} \cdot {\boldsymbol{C}^{r}_\text{match}(x,y)}.
\end{equation}

The hole masks for the reference frames are also aggregated in a similar fashion, resulting in $\boldsymbol{C}_\text{mask}$.
$\boldsymbol{C}_\text{mask}$ indicates pixels that is never visible throughout the reference frame. 

The process of the aggregation is expressed as:

\begin{equation}
    \boldsymbol{C}_\text{mask}(x,y) =
    1 -  ({\sum_{r} {\boldsymbol{C}^{r}_\text{match}({x,y})}}).
\end{equation}

\paragraph{Decoder} 
The decoder network completes the target frame given target features, aggregated reference features, and mask $C_\text{mask}$.
The inputs are concatenated before being fed into the decoder.
Decoder is basically our paste network that learns to fill the missing region by using the aggregated reference features and the visibility of those features. 
The pixels marked on $C_\text{mask}$ are pixels that are never visible in all reference frame because those pixels always fall into holes.
Therefore, the decoder has to be able to synthesize contents for those pixels as well.
We add dilated convolution blocks to grow the receptive field and design the decoder network deeper than the other networks, in order to enhance the completion results for the unseen area by looking at other pixels within the image itself.  

\subsection{Temporal Consistency} 
Each frame in the video is sequentially completed by the network, one by one.
The completed frame at each iteration replaces its reference, providing more information for the following frames as the holes are now filled with contents.
This iterative reference update procedure not only improves the quality of the restored images, but also enhances the temporal consistency.
This is analyzed later in the ablation study.
To further ensure the temporal consistency, we actually run the feed-forward network twice -- completing the video from the first to the last frame, and also in the reverse order. 
Then the final results are computed as follows:
\begin{equation}
\begin{aligned}
    {\hat{\boldsymbol{Y}}^{t}_\text{final}} = \hat{\boldsymbol{Y}}^{t}_\text{forward} \cdot {t\over{N}} + {\hat{\boldsymbol{Y}}^{t}_\text{reverse}} \cdot {(N-t)\over{N}}.
    \label{eq:temporalconsistency}
\end{aligned}
\end{equation}

\section{Training}
\subsection{Loss functions}
All the networks are trained jointly in an end-to-end manner.
First, we compute the loss between the completed target frame and the ground truth. 
The losses for the hole region and the non-hole region are separately calculated.
Furthermore, the hole region can be divided into areas depending on whether the pixel value can be copied from reference frames or not. Therefore, we also apply the losses in the hole region separately.
\begin{equation}
\begin{aligned}
    \mathcal{L}_{\text{hole(visible)}} &= \sum_{t}^{N} {\boldsymbol{M}^{t}} \odot \boldsymbol{C}_\text{mask}\odot||{\hat{\boldsymbol{Y}}^t - \boldsymbol{Y}^t}||_{1},\\
    \mathcal{L}_{\text{hole(invisible)}} &= \sum_{t}^{N} {\boldsymbol{M}^{t}} \odot (1. - \boldsymbol{C}_\text{mask})\odot||{\hat{\boldsymbol{Y}}^t - \boldsymbol{Y}^t}||_{1},\\
    \mathcal{L}_{\text{non-hole}} &= \sum_{t}^{N} {(1- \boldsymbol{M}^{t}}) \odot||{\hat{\boldsymbol{Y}}^{t} - \boldsymbol{Y}^{t}}||_{1}.
    \label{eq:lossinout}
\end{aligned}
\end{equation}
$\boldsymbol{C}_{\text{mask}}$ is properly resized to fit the size of the target frame.

To further improve the visual quality of the results, we also apply perceptual, style, and total variation loss.

\begin{equation}
\begin{aligned}
    \mathcal{L}_{\text{perceptual}} &= {1\over{P}} \cdot \sum_{p}^{P}||{{\phi_{p}({\hat{\boldsymbol{Y}}_\text{comp})}} - \phi_{p}({\boldsymbol{Y}})}||_{1},\\
    \mathcal{L}_{\text{style}} &={1\over{P}} \cdot \sum_{p}^{P}||{{G^\phi_{p} ({\hat{\boldsymbol{Y}}_\text{comp})}} - G^\phi_{p}({\boldsymbol{Y}})}||_{1},\\
    \label{eq:perceptualstyle}
\end{aligned}
\vspace{-10pt}
\end{equation}
where $\hat{\boldsymbol{Y}}_{\text{comp}}$ is combination of the decoder output $\hat{\boldsymbol{Y}}^{t}$ in the hole region and the input $\boldsymbol{X}^{t}$ outside the hole, 
$\phi$ is the output of the pooling layer in pretrained VGG-16~\cite{simonyan2014very} on ImageNet~\cite{imagenet2009cvpr}, $p$ is the pooling index, $G$ is the gram matrix multiplication~\cite{johnson2016perceptual}.

The total-loss function is as follows:
\begin{equation}
\begin{aligned}
    \mathcal{L} &= 2\cdot\mathcal{L}_{\text{align}} + 10\cdot \mathcal{L}_{\text{hole(visible)}} + 20\cdot \mathcal{L}_{\text{hole(invisible)}} \\&+
    6\cdot\mathcal{L}_{\text{non-hole}} + 0.01\cdot \mathcal{L}_{\text{perceptual}} + 24\cdot\mathcal{L}_{\text{style}} + 0.1\cdot\mathcal{L}_{\text{tv}},
\end{aligned}
\end{equation}
where $\mathcal{L}_{\text{tv}}$ is the total variation loss for smoothing the checkerboard effect~\cite{johnson2016perceptual}.
The weight for each loss is empirically determined. 



\begin{figure}
\centering
\includegraphics[width=1.0\linewidth]{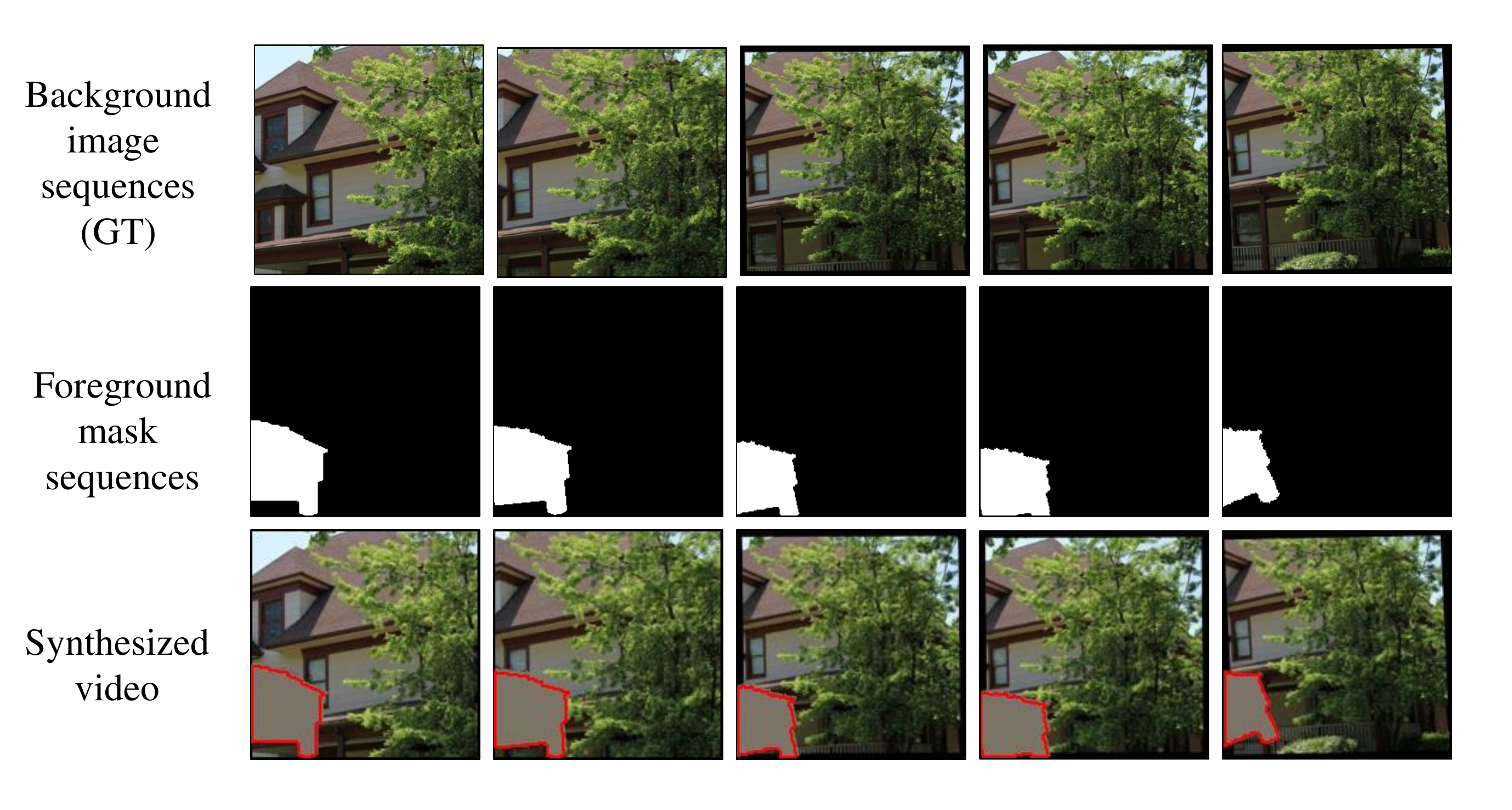}
\caption{Synthesized training dataset example.}
\vspace{-10pt}
\label{fig:Datasets}
\end{figure}
\subsection{Datasets}

Our goal is to complete holes in video sequences.
Inputs are image sequences with holes and binary masks indicating the hole regions.
However, no public video dataset for video inpainting exist.
Therefore, we synthesized a dataset for video inpainting using background images and segmentation masks.

We synthesize videos by compositing background image sequences with object masks (\fref{fig:Datasets}).
To build background image sequences, we use the Places (amount of 1.8M images)~\cite{Places2017TPAMI} single image datasets. To synthesize a sequence of images from a single image, we applied random crops and successive random transformations (shear, scale, translation, rotation) on the image. 
Additionally, we crawled the Youtube video clips and divided them according to the scene (7.3K scenes). Frames are randomly sampled from video clips to form a image sequence. The source of the background image sequence is randomly selected in an equal chance.

To simulate masks for holes, we use object masks from MIT Saliency Benchmark(amount of 11K masks)~\cite{mit-saliency-benchmark} and Pascal VOC 2012(amount of 14.3K masks)~\cite{pascal-voc-2012}.
A mask is randomly resized to be smaller than the size of the background frames.
And the mask is randomly transformed to be a mask sequence by simulating the moving objects.
A training sample is made by compositing a background image sequence and a mask sequence made above. 

\subsection{Training Details}
Our model runs on hardware with the Intel(R) Core(TM) i7-7800X CPU(3.50GHz) CPU and NVIDIA TITAN XP GPUs. We train with the randomly selected five $256\times256$ frames from the synthesized video sequences as inputs.
To train the network, we set the batch size as 40. We use the Adam Optimizer~\cite{kingma2014adam} with learning rates $10^{-4}$ and reduce the running rate factor of 10 every 1 million iterations. The training process takes about 7 days using three NVIDIA TITAN XP GPUs.

\section{Experiments}
To evaluate our algorithm, we provide both quantitative and qualitative analysis, as well as a user study. 
{We conducted the experiments using the videos, which were scaled in half ($424 \times 240$).}
Our code will be available online.
We also show an application of our work in restoring under/over-exposed images. 

\subsection{Quantitative Results}
We first conducted quantitative evaluation by measuring the quality of video restoration.
For this experiment, we randomly selected 25 video sequences in DAVIS dataset ~\cite{DAVIS2016Arxiv, DAVIS2017Arxiv},
which consists of pairs of video and object segmentation mask sequences.
To simulate image restoration, we synthesized videos by putting imaginary object masks from DAVIS~\cite{DAVIS2016Arxiv, DAVIS2017Arxiv} on the videos.
The video without the object masks are used as the ground truth. 
\Tref{table:quantitative results} compares the PSNR and the SSIM measures between our method and~\cite{Huang2016siggraph}.
Both methods show good performance with similar measures. 
Note that VINet~\cite{Kim2019CVPR} is excluded in this experiment because the official code has not been published yet.

\begin{table}
\begin{center}
\begin{tabular}{|l|c|c|}
\hline
Method & PSNR & SSIM  \\
\hline\hline
Huang~\etal~\cite{Huang2016siggraph}  & 28.14  & 0.859 \\
Ours & 28.37 & 0.851\\
\hline
\end{tabular}
\end{center}
\caption{Quantitative Results (video restoration) for DAVIS 2017}
\vspace{-10pt}
\label{table:quantitative results}
\end{table}



\begin{figure*}
\centering
\includegraphics[width=1.0\linewidth]{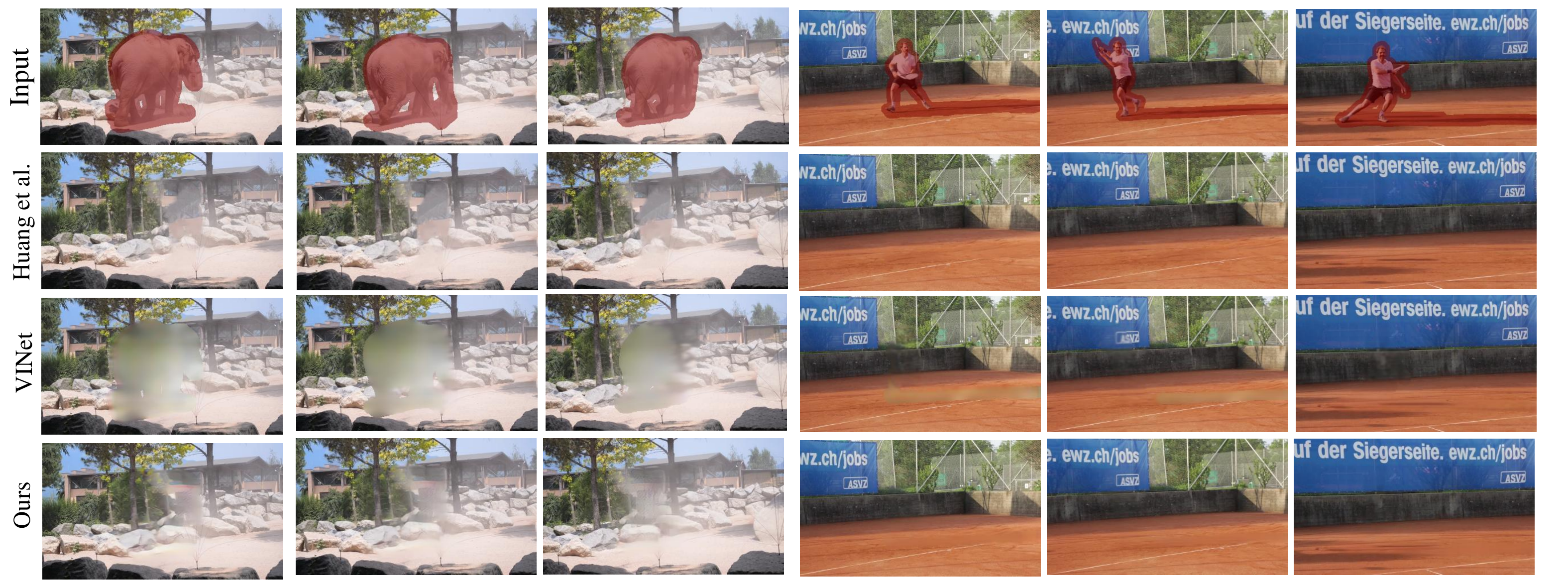}
\caption{
{Qualitative comparison of object removal results for the scenes \textit{elephant} (left) and \textit{tennis} (right) from DAVIS 2017 sequences. }
}
\label{fig:qualitative results}
\end{figure*}
    
\begin{figure*}
\centering
\includegraphics[width=1.0\linewidth]{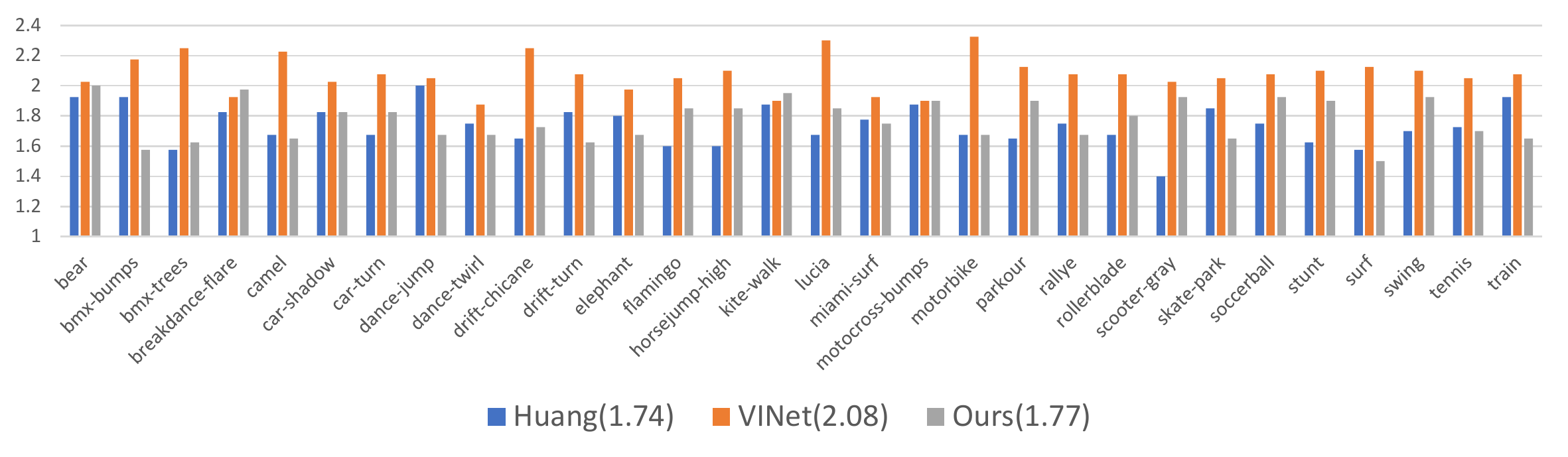}
\caption{User study {for video object removal }results (lower value is better)}
\label{fig:userstudy}
\end{figure*}

\subsection{User Study and Qualitative Analysis}
We further conducted experiments on dynamic object removal in videos with 30 videos from DAVIS dataset~\cite{DAVIS2016Arxiv,DAVIS2017Arxiv}. 
We compared our methods with the state-of-the-art video inpainting models~\cite{Huang2016siggraph, Kim2019CVPR}.
Results of the previous methods were gathered by using the official code released by the authors~\cite{Huang2016siggraph} and by requesting the results from the authors~\cite{Kim2019CVPR}.

The user study result performed the Amazon Mechanical Turk (AMT) is shown in \fref{fig:userstudy} and \Tref{table:userstudy} . 
The workers were asked to rank the video completion results and we also allowed them to give ties.
All tests were evaluated by 40 participants. 
\begin{table}[h]
\begin{center}
\begin{tabular}{|l|c|}
\hline
Method & Average ranking  \\
\hline\hline
Huang~\etal~\cite{Huang2016siggraph}  & 1.74 \\
VINet~\cite{Kim2019CVPR} & 2.08\\
Ours & 1.77 \\
\hline
\end{tabular}
\end{center}
\caption{User study average rank (lower value is better)}
\vspace{-10pt}
\label{table:userstudy}
\end{table}

The user study shows that our method is highly competitive to the optimization based method~\cite{Huang2016siggraph}, while VINet~\cite{Kim2019CVPR} is not on par with the other two methods. 
While the method in \cite{Huang2016siggraph} was slightly more favored, it requires average completion time of 952 seconds per video, whereas our method only takes \textbf{27.14} seconds. 

Qualitative comparisons of the object removal results are shown in \fref{fig:qualitative results}.
These comparisons show similar results as the user study.
Our results are comparable to the state-of-the-art method in \cite{Huang2016siggraph}, while showing much better results compared to the other deep learning based approach in \cite{Kim2019CVPR}.

\subsection{Applications}
We extend our method for restoring under/over-exposed image sequences. The restoration process is similar to video inpainting problem in that it fills areas with missing information. 
This problem often happens to image sequences taken by a camera attached to a vehicle due to rapid exposure changes (\eg tunnel entry and exit). 


\begin{figure*}
\centering
\includegraphics[width=1.0\linewidth]{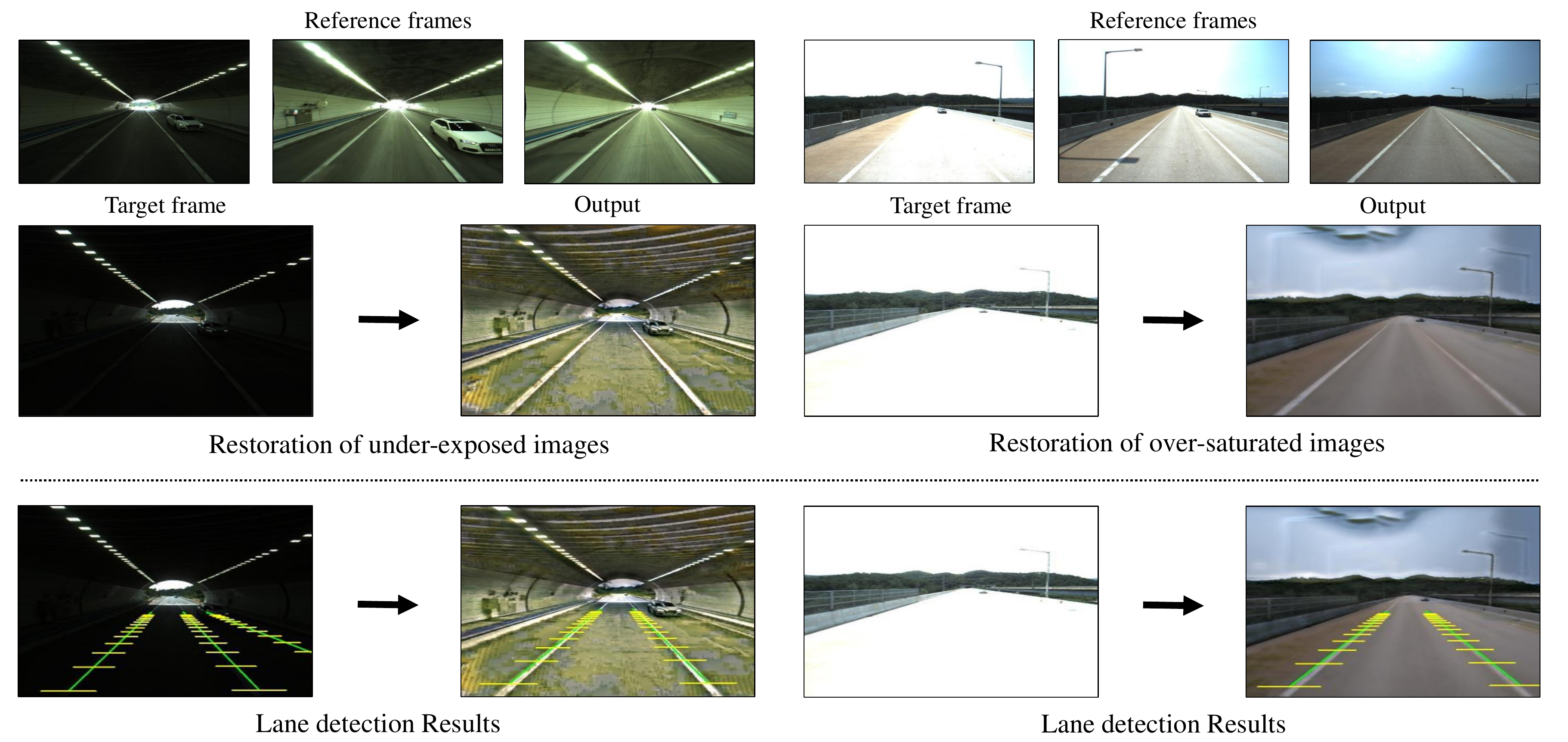}
\caption{Application of our method for the restoration of under/over-exposed images.}
\vspace{-10pt}
\label{fig:application}
\end{figure*}

As shown in \fref{fig:application}, both the texture and the color are improved. 
To validate the effectiveness of our restoration process, we ran a lane detection algorithm on road images before and after the enhancement.  
We collected 469 frames videos
\footnote{{The dataset were taken by using Mobile Mapping System Camera of Hyundai MnSOFT, Inc.}} 
that contains rapid exposure changes due to tunnels and the internal color histogram-based lane detection method was used.
As shown in \fref{fig:application} and \Tref{table:lanedetectionaccuracy}, lane detection results are significantly improved. 
\begin{table}
\begin{center}
\begin{tabular}{|l|c|}
\hline
Lane detection input & Lane detection accuracy  \\
\hline\hline
Over/under-exposed image  & 46.69\% \\
Restored input by our model&  \textbf{83.00\%}\\
\hline
\end{tabular}
\end{center}
\caption{The lane detection accuracy results. }
\label{table:lanedetectionaccuracy}
\end{table}

\begin{figure}
\centering
\includegraphics[width=1.0\linewidth]{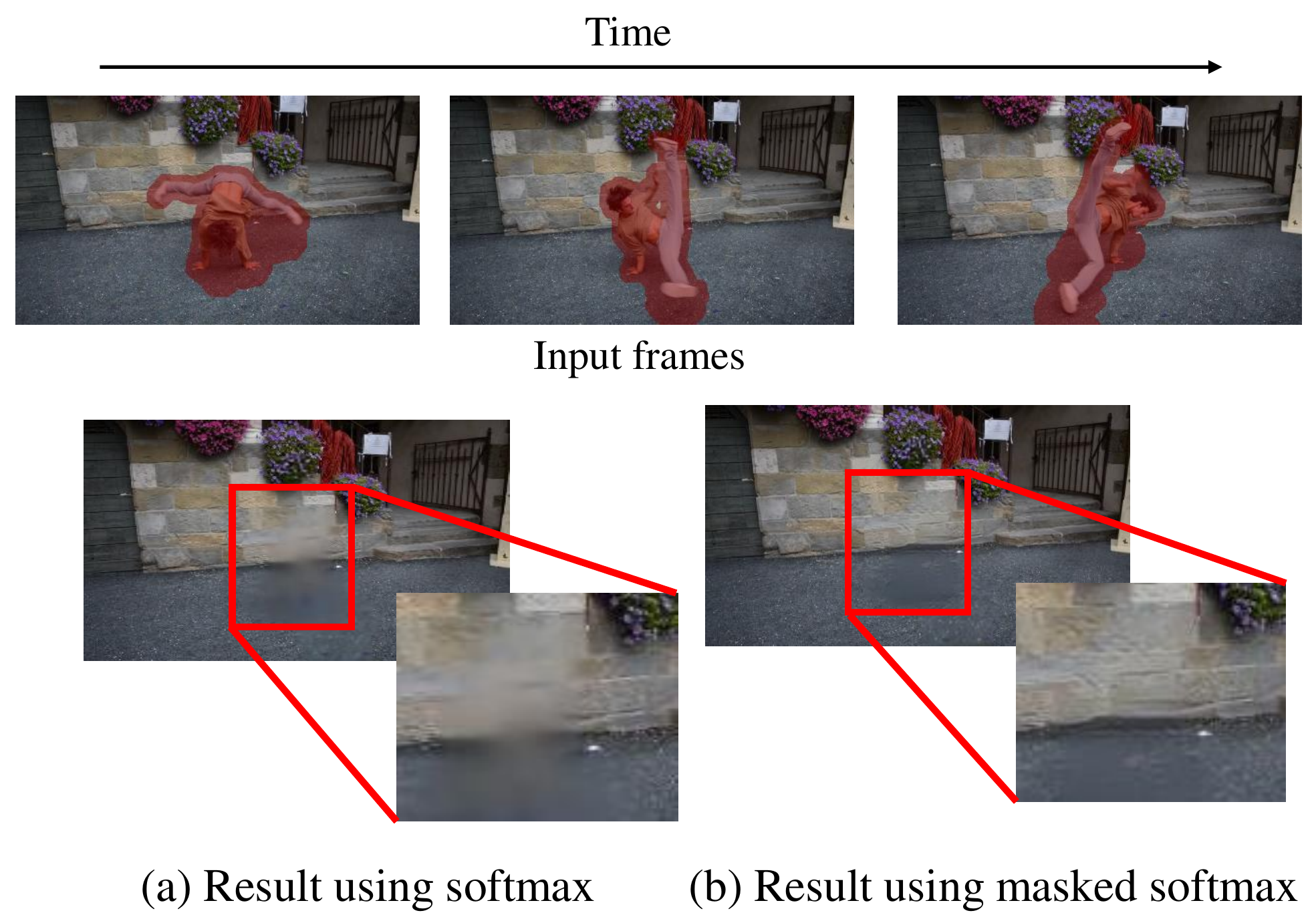}
\caption{Ablation study for masked softmax.}
\vspace{-5pt}
\label{fig:ablationstudy}
\end{figure}

\section{Ablation Study}
\paragraph{Masked softmax} We conducted an ablation study to verify that masked softmax contributes to the performance improvements. We train our model using normal softmax under the same conditions. As shown in the \fref{fig:ablationstudy}, using masked softmax results are sharper than using the normal one. 

\begin{figure}
\centering
\includegraphics[width=1.0\linewidth]{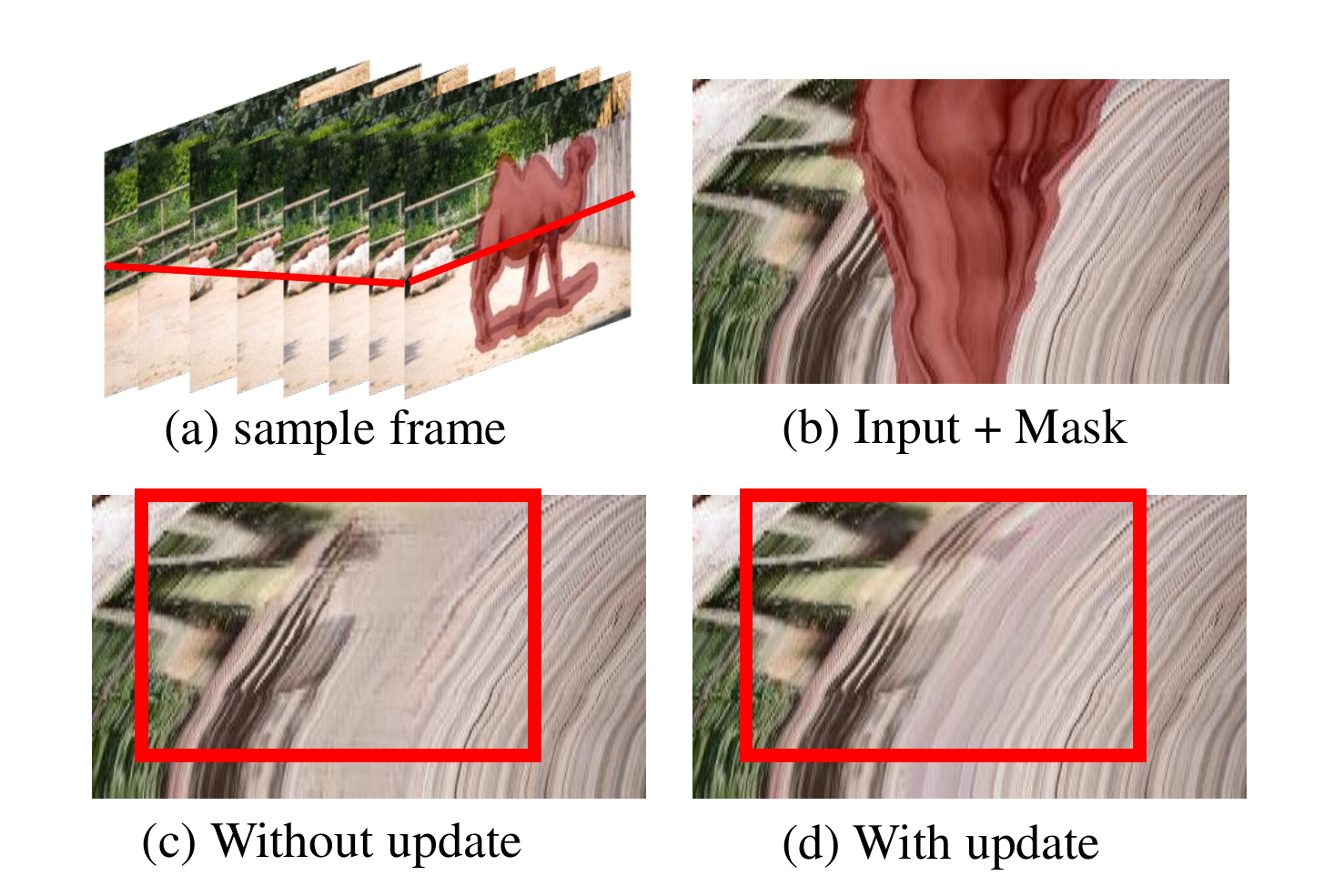}
\caption{Ablation study for reference update. 
{(b), (c) and (d) show the temporal profile of the red line shown in input (a).} 
}
\vspace{-5pt}
\label{fig:TemporalCoherent}
\end{figure}

\paragraph{Reference update}
To produce temporally coherent outputs, we update the past reference frames with the inpainted version.   
To visualize the effect of this updating protocol, we compare the temporal profile~\cite{caballero2017real} of resulting videos in \fref{fig:TemporalCoherent}.
As shown in \fref{fig:TemporalCoherent}, the update procedure contributes in enhancing the temporal consistency. 

\section{Conclusion}
In this paper, we presented a novel DNN framework for video inpainting. The proposed method inpaints the missing information by copy-and-pasting contents from the reference frames. 
The reference information is dynamically updated by the previous completion results to ensure the temporal consistency. 
Our experiments support that the proposed framework is comparable to the optimization-based methods and outperform other deep learning based approaches. We extended our framework to restore over/under-exposed in videos and were able to significantly increase the lane detection accuracy.

\section*{Acknowledgement}\noindent
This work was supported by Institute for Information \& communication Technology Promotion  (IITP) grant funded by the Korea government (MSIP) (2018-0-01858).
{\small
\bibliographystyle{ieee_fullname}
\bibliography{egbib}
}

\end{document}